\documentclass[11pt]{article}
\usepackage{eamt23}
\usepackage{times}
\usepackage{url}
\usepackage{latexsym}
\usepackage{amssymb}
\usepackage{amsmath}
\usepackage[small,bf]{caption} % added MLF 20171211
\usepackage{tabularx} % in preamble

\usepackage{subcaption}
\usepackage[T1]{fontenc}
\usepackage{float}
\makeatletter
\providecommand{\@internalcopyrightfn}{}
\makeatother
\usepackage{hyperref} 
\usepackage{booktabs}
\usepackage{tcolorbox}

\hypersetup{
    colorlinks=true,
    linkcolor=[rgb]{0,0,0.6},   % #000099
    urlcolor=[rgb]{0,0,0.6},
    citecolor=[rgb]{0,0,0.6},
    pdfborder={0 0 0}           % removes boxes
}

\title{\texttt{ForMaT}: Dataset for Visually-Grounded Multilingual PDF Translation}

\author{
Michał Ciesiółka$\normalfont{\textsuperscript{1,2}}$, 
Dawid Wiśniewski$\normalfont{\textsuperscript{1,3}}$, 
Adrian Charkiewicz$\normalfont{\textsuperscript{1}}$, 
Kamil Guttmann$\normalfont{\textsuperscript{1,2}}$ \\
$\textsuperscript{1}$ Laniqo, Poznań, Poland \\
$\textsuperscript{2}$ Faculty of Mathematics and Computer Science, Adam Mickiewicz University, Poznań, Poland \\
$\textsuperscript{3}$ Poznań University of Technology \\
{\tt \{name\}.\{surname\}@laniqo.com}
}
\date{}

\begin{document}
\maketitle
\begin{abstract}
We present \texttt{ForMaT} (\textbf{For}mat-Preserving \textbf{M}ultilingu\textbf{a}l \textbf{T}ranslation), a parallel corpus of 3,956 PDFs across 15 language pairs that preserves original layout metadata proposed for multimodal machine translation. To ensure structural diversity in the dataset, we employ K-Medoids sampling over 45 geometric features, capturing complex elements like nested tables and formulas to focus only on visually diverse PDF documents. Our evaluation reveals that current MT systems struggle with spatial grounding and geometric synchronization, often losing the link between text and its visual context. \texttt{ForMaT} provides a benchmark for developing layout-aware translation models that integrate visual and textual context for high-fidelity document reconstruction.
\end{abstract}

\section{Introduction}
Modern machine translation (MT) systems increasingly leverage multimodal signals to enhance translation accuracy. In the audio domain, for instance, paralinguistic features, such as speaker identification, gender, and emotional tone, provide essential context that helps disambiguate intent and refine target-language nuances. Similarly, when processing visually rich documents like PDF files, visual and spatial cues are often indispensable for resolving lexical ambiguity and selecting the appropriate morphological forms. This shift toward context-dependent, multimodal translation has emerged as a significant frontier in MT research~\cite{shen2024surveymultimodalmachinetranslation,DBLP:conf/emnlp/FengLHHN25}.

In this paper, we focus on visual context through the introduction of a new parallel corpus comprising 3,956 PDF documents, which we named \texttt{ForMaT} (\textbf{For}mat-Preserving \textbf{M}ultilingu\textbf{a}l \textbf{T}ranslation). The dataset spans 15 language pairs involving English, German, Spanish, French, Italian, and Polish. Unlike traditional parallel corpora that are limited to plain text, our dataset preserves the rich layout and formatting information of the original source documents.

Our contributions are threefold: first, we motivate the necessity of layout-aware resources for modern MT; second, we detail a methodology for dataset collection and provide an in-depth analysis of the corpus’s structural properties; and third, we establish the dataset’s diversity through a multi-dimensional analysis, positioning it as a benchmark for evaluating the next generation of layout-aware and document-level translation systems. To demonstrate the practical utility of this benchmark, we conclude by evaluating several state-of-the-art PDF translation systems, specifically analyzing their ability to preserve both linguistic meaning and complex document layouts.

\subsection{Motivation}
In machine translation, visual cues within a source document are often essential for generating accurate target output. When processing PDF files, several use cases demonstrate why visual and spatial context is critical:
\begin{itemize}
    \item Image captions -- Visual context helps disambiguate polysemic words and personal pronouns. For instance, an image depicting a woman can signal the use of the feminine pronoun \textit{she} when translating from gender-neutral languages (e.g., Turkish, Hungarian, or Basque). Similarly, short captions often lack sufficient textual context; an image can clarify whether the word \textit{head} in a medical document refers to an anatomical body part, a team leader, or the top element of a device.
    \item Text position -- Spatial positioning helps identify named entities and document semantics. In an invoice, for example, a company name is typically expected at the top, while numerical values in specific regions are more likely to represent monetary amounts rather than dates or quantities.
    \item Tables -- The translation of table cells often depends on context provided by adjacent cells or headers. Depending on the layout, these headers may appear in the top rows (column-wise representation) or the flanking columns (row-wise representation). Such structures pose unique challenges \cite{yin-etal-2020-tabert}, as the visual layout of a PDF helps a model identify that a text fragment is part of a table, allowing it to focus on the correct relational context to understand cell content.
    \item Text Segmentation -- The spacing between words and paragraphs is a vital indicator of context. Traditional OCR tools paired with MT models often fail when text is justified with non-standard spacing, written in creative or non-linear formats (e.g., one word per line), or rotated vertically. Identifying text clusters through visual analysis allows the model to segment the document into meaningful units, preserving the intended translation context. 
    \item Geometric Constraints -- Document layout serves as a guide for translation length and copy-fitting. Because target languages may vary significantly in word length, the layout dictates how the translation must be aligned. By analyzing the visual properties of the PDF, models can select translations that better fit the available space, minimizing unnatural gaps or managing page breaks more effectively.
\end{itemize}

To address these challenges, MT datasets must evolve to include images and rich formatting (such as original PDFs). This integration enables a more robust evaluation of how translation models perform when document structure is inextricably linked to meaning.

\section{Related works}

The evolution of Machine Translation (MT) has been marked by a steady expansion of the context window, moving from isolated sentences to entire documents and, more recently, to multimodal inputs. \texttt{ForMaT} sits at the intersection of Document-level MT, Multimodal Learning, and Visually-Rich Document Understanding (VRDU).
\subsection{Multimodal and Document-Level MT}

Early efforts in Multimodal Machine Translation (MMT) focused primarily on visual grounding for image captioning, exemplified by the Multi30K~\cite{DBLP:conf/acl/ElliottFSS16} dataset. These tasks used images to resolve lexical ambiguities (e.g., gender or entity type) but were limited to short, isolated sentences. Recent research has shifted toward a more context dependent approaches utilizing various modalities, where researchers leverage non-textual signals to improve translation quality in specific domains~\cite{shen2024surveymultimodalmachinetranslation,DBLP:conf/emnlp/FengLHHN25}.

While Document-Level MT (DMT) addressed long-range textual dependencies, most existing benchmarks, including the massive DocHPLT~\cite{DBLP:journals/corr/abs-2508-13079}, rely on "flattened" web-crawled text. These corpora lack the 2D spatial information (headers, footers, sidebars) that is vital for interpreting the logical flow of high-stakes documents like technical manuals or legal acts.

\subsection{Visually-Rich Document Understanding (VRDU)}

The field of VRDU has established that spatial coordinates and typographic cues are as important as the text itself for understanding complex layouts. The LayoutLM series (v1, v2, v3)~\cite{huang2022layoutlmv3pretrainingdocumentai} pioneered the use of 2D positional embeddings to model the relationship between text and visual structure. More recently, Layout-Aware LLMs have demonstrated that encoding document geometry as specialized tokens can significantly improve performance in information extraction and Document Visual Question Answering (VQA)~\cite{lu-etal-2025-bounding}.

Other popular models aimed at VRDU task are: TaBERT~\cite{yin-etal-2020-tabert}, focused on the joint understanding of textual and tabular data, or DocLayout-YOLO~\cite{zhao2024doclayoutyoloenhancingdocumentlayout}, PP-DocLayout~\cite{Sun2025-mg}, and the PaddleOCR 3.0~\cite{Cui2025-sd} all focused on layout understanding.

However, a gap remains: while VRDU models excel at extraction, they are rarely evaluated on generative translation. \texttt{ForMaT} provides the necessary parallel data to bridge this gap, treating translation not just as a linguistic task, but as a layout-preservation task. %While other 

\subsection{Multimodal LLMs}
Modern approaches leverage Multimodal Large Language Models (MLLMs) -- language models with the ability to understand modalities other than texts~\cite{10.1093/nsr/nwae403}. Recent models that focus on images, introduce various optimizations helping understand different modalities, e.g., utilizing visual instruction tuning~\cite{NEURIPS2023_6dcf277e} and global-local dual perception~\cite{lu2026globallocaldualperceptionmllms} for high-resolution images. Recent systems like InImageTrans~\cite{inimagetrans}, TranslateGemma~\cite{finkelstein2026translategemma}, or Gemini~\cite{geminiteam2025geminifamilyhighlycapable} demonstrate the potential of LLMs to handle "visually-situated" text and translate between languages. Furthermore, research into zero-shot MMT~\cite{futeral-etal-2025-towards} and unimodal alignment~\cite{DBLP:conf/cvpr/ZhangYA25} aims to reduce the reliance on costly supervised parallel data.

\subsection{Benchmarks for Document Image Translation (DIMT)}

The most recent frontier in the field is Document Image Machine Translation (DIMT), highlighted by the ICDAR 2025 DIMT Challenge~\cite{zhang2025icdar}. Current state-of-the-art benchmarks such as DIMT-WebDoc-300K and DIMT-arXiv-124K focus heavily on translating document images into Chinese, often prioritizing scale over structural variety.

Similarly, while M3T~\cite{DBLP:conf/naacl/HsuLLFNNLKP24} has introduced document-level multimodal machine translation benchmarks, there remains a need for a corpus sampled specifically for structural difficulty. \texttt{ForMaT} addresses this gap by using a rigorous K-Medoids sampling~\cite{kaufman1990kmedoids} methodology across 45 structural features to collect only diverse PDF documents, ensuring that the corpus serves as a stress test for the next generation of layout-aware translation models.

\subsection{Evaluating Multimodal Models}
While traditional metrics remain standard, recent findings by~\cite{sun-etal-2025-fine} indicate that n-gram overlap scores like BLEU often fail to capture document-level coherence, advocating for more nuanced, multi-dimensional evaluation. This shift is reflected in the ICDAR 2025 DIMT Challenge~\cite{zhang2025icdar}, which underscores the persistent difficulty of translating complex layouts. Methods, such as multimodal reasoning and dual-perception architectures try to address this challenge~\cite{huang2026step3,lu2026globallocaldualperceptionmllms}.
%This challenge that models like Step3~\cite{huang2026step3} attempt to solve via multimodal reasoning and dual-perception architectures~\cite{lu2026globallocaldualperceptionmllms}.

In comparison to existing literature, \texttt{ForMaT} offers three distinct advantages:

\begin{enumerate}
\item \textbf{Language Diversity}: We provide 15 language pairs, with a focus on European languages often underrepresented in recent DIMT challenges.

\item \textbf{Structural Complexity}: Unlike datasets that rely on random crawling, we employ K-Medoids sampling over 45 structural features to ensure our corpus includes challenging cases like complex tables, inline formulas, multiple columns, and images with captions.

\item \textbf{High-Fidelity Metadata}: We provide raw layout metadata alongside the parallel text, enabling the development of models that can reconstruct the target PDF with pixel-perfect accuracy.
\end{enumerate}

\section{Dataset}

\begin{figure}[t]
    \centering
    \includegraphics[width=0.45\columnwidth]{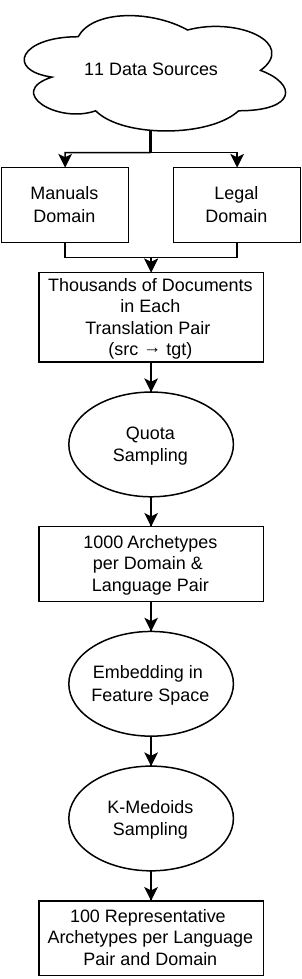}
    \caption{ForMaT dataset collection process. Each operation was performed independently for each language pair in both domains.}
    \label{fig:data_flow}
\end{figure}

The process of collecting the dataset consists of several phases as depicted in Figure~\ref{fig:data_flow}. First, we identify websites providing PDF documents that meet our selection criteria, then, we sample a first, broad collection of documents using quota sampling. Finally, we filter the broad collection of documents, to leave only the most diverse and interesting documents in terms of visual complexity and composition. The final \texttt{ForMaT} dataset is represented by 3,956 documents.

\subsection{Data sources}

To create our dataset, we targeted two domains 
exhibiting distinct linguistic profiles and visual formats:
legal documents and technical user manuals. 

For the legal domain, we curated a corpus from international and national institutions that publish official documentation in a multilingual format. A significant portion of this data was sourced from European Union repositories, specifically focusing on three distinct legal contexts: legislative acts via EUR-Lex, parliamentary proceedings through the European Parliament portal, and judicial documentation from the European e-Justice Portal. To ensure broader geographic and administrative variety, we further incorporated the corpus with documents from the United Nations digital library, the Swiss federal law repository (Fedlex), and the U.S. Social Security Administration (SSA).

To curate the user manual domain, we utilized a global index of electronics brands\footnote{\url{https://en.wikipedia.org/wiki/List_of_electronics_brands}} as a foundational blueprint for our search. To facilitate downstream document processing and alignment, we exclusively selected manufacturers that publish localized instructions as individual, single-language files rather than consolidated multilingual volumes. Although our initial search targeted electronic product domain, we expanded the scope to include major automotive manufacturers to increase the variety of instructional layouts. The final selection included documentation from Huawei, Lenovo, Philips, Nissan and Toyota, providing a diverse set of technical terminologies and visual schematics. The full set of sources for both domains is summarized in Table \ref{tab:sources}.

% During the data collection process, a portion of the initially identified documents was excluded due to licensing or terms-of-use restrictions that prevented automated collection or redistribution and was not mentioned in the Table \ref{tab:sources}. Consequently, the final dataset contains only documents whose usage conditions permit research use.

The choice of the domains and data sources depended on the licenses assigned to the documents; we collected only those documents that can be used for research purposes.

\begin{table}[ht]
\centering
\small
\ttfamily
\caption{Dataset Sources and URLs}
\label{tab:sources}
\begin{tabularx}{\columnwidth}{l>{\raggedright\arraybackslash}X}
\hline
\textbf{Source} & \textbf{URL} \\
\hline
United Nations & \href{https://www.un.org/en/}{un.org} \\
SSA & \href{https://www.ssa.gov/}{ssa.gov} \\
E-Justice & \href{https://e-justice.europa.eu}{e-justice.europa.eu} \\
Fedlex & \href{https://www.fedlex.ch/}{fedlex.ch} \\
European Parliament & \href{https://www.europarl.europa.eu/}{europarl.europa.eu} \\
EUR-Lex & \href{https://eur-lex.europa.eu/}{eur-lex.europa.eu} \\
Toyota & \href{https://www.toyota-europe.com/customer/manuals}{toyota-europe.com} \\
Philips & \href{https://www.usa.philips.com/healthcare/support/resource-center}{philips.com} \\
Nissan & \href{https://www.nissan-techinfo.com}{nissan-techinfo.com} \\
Lenovo & \href{https://pcsupport.lenovo.com/}{lenovo.com} \\
Huawei & \href{https://consumer.huawei.com}{huawei.com} \\
\hline
\end{tabularx}
\end{table}

For each domain, we have collected only parallel data available in multiple languages, where source and target documents are expressed in: English, French, German, Italian, Polish, and Spanish. This set of languages was found representative, as the attempts to add new languages to the set resulted in unrepresented language pairs at the sampling stage.

% Initially, we considered additional languages, but each new language substantially increased the number of language pairs, which created underrepresented categories. We therefore focused on this subset to ensure that all language pairs were equally and fully represented.

\subsection{Data sampling}

To balance data across the two primary domains and fifteen language pairs, we targeted a sample of 1,000 documents per pair in each domain.

We adopted a quota sampling strategy \cite{cochran1977sampling} with two modifications to address imbalances in underrepresented sources. First, we grouped documents by source and language pair, then sampled them in ascending order of available documents. This ensured that underrepresented groups were fully included before larger sources were tapped. Second, when a source failed to meet its quota, the remaining capacity was dynamically redistributed to larger sources.

To capture potential document changes in time, we sampled the EUR-Lex corpus evenly by year. Given its unique volume, we selected two search-result pages per year from 2005 to 2025, specifically choosing sets where every document was available in all six target languages. This approach allows us to observe the evolution of official language and structure over two decades. For each document, we collected and analyzed only the first 10 pages.
% Finally, for long documents across all sources, we limited entity and data extraction to the first 10 pages.

% During the data collection process, a portion of the initially identified documents was excluded due to licensing or terms-of-use restrictions that prevented automated collection or redistribution. Consequently, the final dataset contains only documents whose usage conditions permit research use.

\subsection{Data retrieval}

To ensure the selection of stylistically diverse PDFs, we developed a hybrid extraction pipeline that integrates computer vision for layout detection with low-level PDF parsing for precise text and metadata extraction. This approach allowed us to filter our initial pool of 1,000 documents per domain/language pair, retaining only those with the most diverse structural and formatting features.

\subsubsection{Visual Layout Analysis}
We employed the \texttt{PaddleOCR} \cite{Cui2025-sd} library, specifically the \texttt{PP-DocLayoutV2} \cite{Sun2025-mg} model, to analyze the visual structure of each document. Unlike standard OCR models that prioritize text character recognition, this model treats the PDF page as a visual image to identify high-level semantic regions. It segments the page into discrete categories, including headers, footers, figures, tables, and standard text blocks.

\subsubsection{Textual Extraction and Metadata Parsing}
Complementing the visual analysis, we utilized the \texttt{pdfminer}\footnote{\url{https://github.com/pdfminer/pdfminer.six}} library to extract styling metadata directly from the PDF file. We chose direct parsing over OCR-based recognition to ensure the high-fidelity preservation of formatting attributes.

Our extraction module retrieves word-level styling information, including font family, size, color, and weight. These attributes are critical features for document translation pipelines: they ensure that formatting properties from the source text (e.g., a specific token marked in bold or red) are accurately mapped to the corresponding fragment in the target translation.

\subsubsection{Vectorization}
To quantify the structural complexity of the sampled documents, each file was mapped to a 45-dimensional feature vector $\vec{v}$ (where $\vec{v} \in \mathbb{R}^{45}$), as detailed in Table~\ref{tab:vector_mapping}. These features integrate entity types identified via \texttt{PaddleOCR} with formatting metadata from \texttt{pdfminer}, and are organized into three primary categories:

\begin{itemize}
    \item \textbf{Text-Based Labels:} These features represent the average frequency of textual elements per page. This category includes structural elements such as \texttt{text}, \texttt{footer}, \texttt{paragraph\_title}, and \texttt{abstract}.
    \item \textbf{Visual Labels:} These features capture non-textual or complex graphical entities within the layout, including technical structures (e.g., \texttt{table}, \texttt{algorithm}), graphical assets (e.g., \texttt{image}, \texttt{seal}), and specialized spatial formatting such as \texttt{vertical\_text}.
    
    \item \textbf{Typographic and Stylistic Attributes:} 
    This category captures the visual properties of the text using a combination of frequency and occurrence metrics. We record the total number of unique font weights (e.g., \texttt{bold}, \texttt{italic}) and distinct font names present in the document. To normalize variations, font sizes are rounded to the nearest 0.5 points before counting. The color profile is represented as a binary sub-vector, where specific indices (e.g., \texttt{blue}, \texttt{black}) are assigned a value of 1 if the color is present in the document text and 0 otherwise.
\end{itemize}

%The full feature mapping is shown .

\subsubsection{Clustering}

% To ensure the final dataset captures the maximum structural and stylistic diversity, we employed the K-Medoids clustering algorithm. Unlike K-Means, which calculates a synthetic "mean" center, K-Medoids selects an actual document from the dataset (the medoid) as the cluster center. This ensures that our selected samples are authentic, high-quality representatives of their respective clusters rather than mathematical averages.

To capture maximum structural and stylistic diversity, we employed the $K$-Medoids clustering algorithm to select representative documents from our vectorized pool. By clustering the documents into $K$ groups and selecting the medoid (the most centrally located document) of each cluster, we ensured that our final selection of $K$ documents represented the full breadth of the data's stylistic variance.

% Using medoids, we selected one actual document from each cluster as a representative.

We performed clustering independently for each language pair within the two domains. To represent a parallel document pair as a single entry, we computed a combined feature representation by averaging the 45-dimensional feature vectors extracted from the source and target documents. These representations were then clustered into $K$=100 distinct groups per language pair in each domain (15 pairs $\times$ 2 domains = 30 processes) using the Euclidean distance metric. To ensure a globally representative selection, we utilized the k-medoids++ initialization strategy, which spreads the initial medoids far apart in the feature space.

This approach reduces the total number of documents while preserving a broad spectrum of complexities, ranging from simple text-only reports to intricate technical schematics. The choice of $K$=100 was a heuristic intended to capture a wide range of layouts without over-fragmentation.

% Following this clustering and selection process, we arrived at a final corpus comprising 1,278 unique document archetypes (identified by \texttt{common\_id}). When expanded across their respective translations, this resulted in a total of 3,956 unique document instances (represented by distinct URLs) across the fifteen language pairs.

In this paper, we refer to the underlying content shared across translations as a document archetype, distinguishing it from a specific PDF file in a specific language, which we call a document instance.
Because many documents in our source pool exist in multiple languages, the independent sampling processes occasionally selected the same document "archetype" as a representative for different language pairs. 
% In this paper, we refer to the underlying content shared across translations as a document archetype, distinguishing it from a specific PDF file in a specific language, which we call a document instance.

As a result of this overlap across the 30 sampling processes, we identified 1,278 unique document archetypes. Since these archetypes do not all exist in every one of the 15 language pairs, we collected all available translations for these specific selections, resulting in a total of 3,956 unique document instances (URLs). This methodology ensures that each language pair is represented by a diverse set of layouts.% while maximizing the utility of available translations.

% even if the "representative" documents are not perfectly symmetrical across the entire corpus.

%\begin{itemize}
%    \item \textbf{Archetype vs. Instance:} A "document archetype" refers to the underlying content shared across translations. A "document instance" refers to a specific PDF file in a specific language.
%    \item \textbf{The final document count:} While the 30 independent samplings could have yielded up to 3,000 selections, the high degree of overlap between language pairs resulted in a set of 1,278 unique document archetypes.
%    \item \textbf{Expansion:} The archetypes do not all exist in every one of the 15 languages. When we gathered all available translations for these specific archetypes, we arrived at a total of 3,956 unique document instances (URLs). This approach ensures that each language pair is represented by diverse layouts, even if the "representative" documents are not perfectly symmetrical across the entire corpus.
%\end{itemize}

\subsubsection{Representativeness and Diversity Gain}

% \textbf{(tables in appendix, might move somewhere in main body)}

Our sampling strategy produced a corpus with significantly greater average structural complexity per document than the initial pool. By selecting cluster medoids rather than random samples, we successfully amplified the presence of underrepresented structural features.

The final corpus is considerably more visually dense than the original dataset. Specifically, the relative frequency of images more than doubled (+101.2\%), while the occurrence of tables increased by 60.9\%. Supporting graphical elements—such as \texttt{vision\_footnotes} and \texttt{figure\_titles} also saw substantial growth, rising by 48\% and 33\%, respectively.

% This confirms that our medoids are structurally coherent, preserving the relationship between visual assets and their associated textual descriptions.

The selection process markedly increased color diversity by capturing stylistic variations typically overlooked by random sampling. While the initial pool was predominantly monochromatic, the final subset exhibited substantial growth in minority colors: the frequencies of Yellow and Red rose by 264\% and 215\%, respectively, while Purple, Teal, and Pink each increased by over 115\%.

The complete impact of the clustering process on entity and color distributions is detailed in Appendix B.

% \subsection{Dataset overview}
% % Describe each field of the dataset
% % [show some numbers, (source and PDF count) (language pairs and count) 

\subsection{Dataset availability}

The dataset is available online:  \href{https://huggingface.co/datasets/laniqo/ForMaT}{Dataset (Hugging Face)}~\footnote{\url{https://huggingface.co/datasets/laniqo/ForMaT}}.

\section{Explorative Data Analysis}

The concept of multidimensional diversity serves as the foundational framework for evaluating the architectural complexity of this corpus. Rather than treating document difficulty as a singular, linear metric, we model it as a coordinate within a multi-axis space defined by structural, and stylistic features. 

Modern translation systems face challenges on both text-level style and layout preservation. This multidimensional approach is essential because document difficulty is rarely uniform; a page might be linguistically simple yet stylistically complex, or visually dense while maintaining a rigid, predictable layout.

\subsection{Feature Independence}

\begin{figure}[ht] % Single column, [ht] tries 'here' then 'top'
\centering
    \includegraphics[width=\columnwidth]{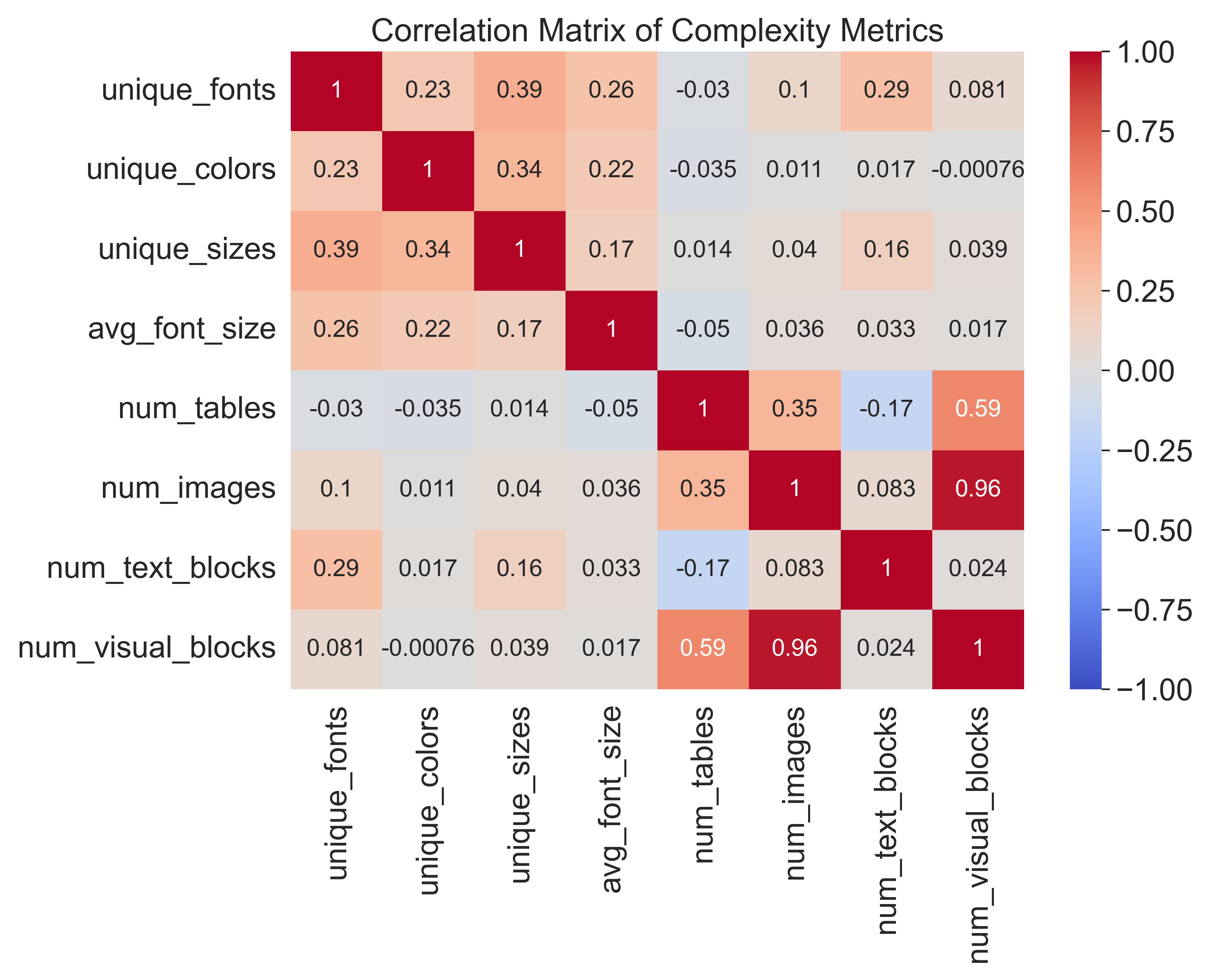}
    \caption{Spearman correlation matrix of document complexity metrics.}
    \label{fig:correlation}
\end{figure}

Figure \ref{fig:correlation} presents a Spearman correlation matrix of selected document attributes. The results indicate low correlation between different dimensions of document variety.
Notably, stylistic attributes (such as the number of unique font colors and sizes) show minimal correlation ($r < 0.2$) with structural indicators like the number of graphical entities. This finding suggests that "visual complexity" (e.g., a page full of images and tables) and "formatting complexity" (e.g., documents with varied font colors and sizes) represent independent challenges.

However, the results also highlight a moderate coupling between typographic variety and textual density. We observe that both the number of unique fonts and the number of unique font sizes show a positive correlation with the number of text blocks ($r = 0.29$ and $r = 0.16$, respectively). This suggests that as a document is divided into more text blocks, the variety of formatting styles increases. This implies that the task of text-level style preservation becomes increasingly difficult in high-density documents, where the system must track a larger volume of independent stylistic metadata alongside the translated content.

\subsection{Structural Variance}

\begin{figure}[ht] % Single column, [ht] tries 'here' then 'top'
\centering
    \includegraphics[width=\columnwidth]{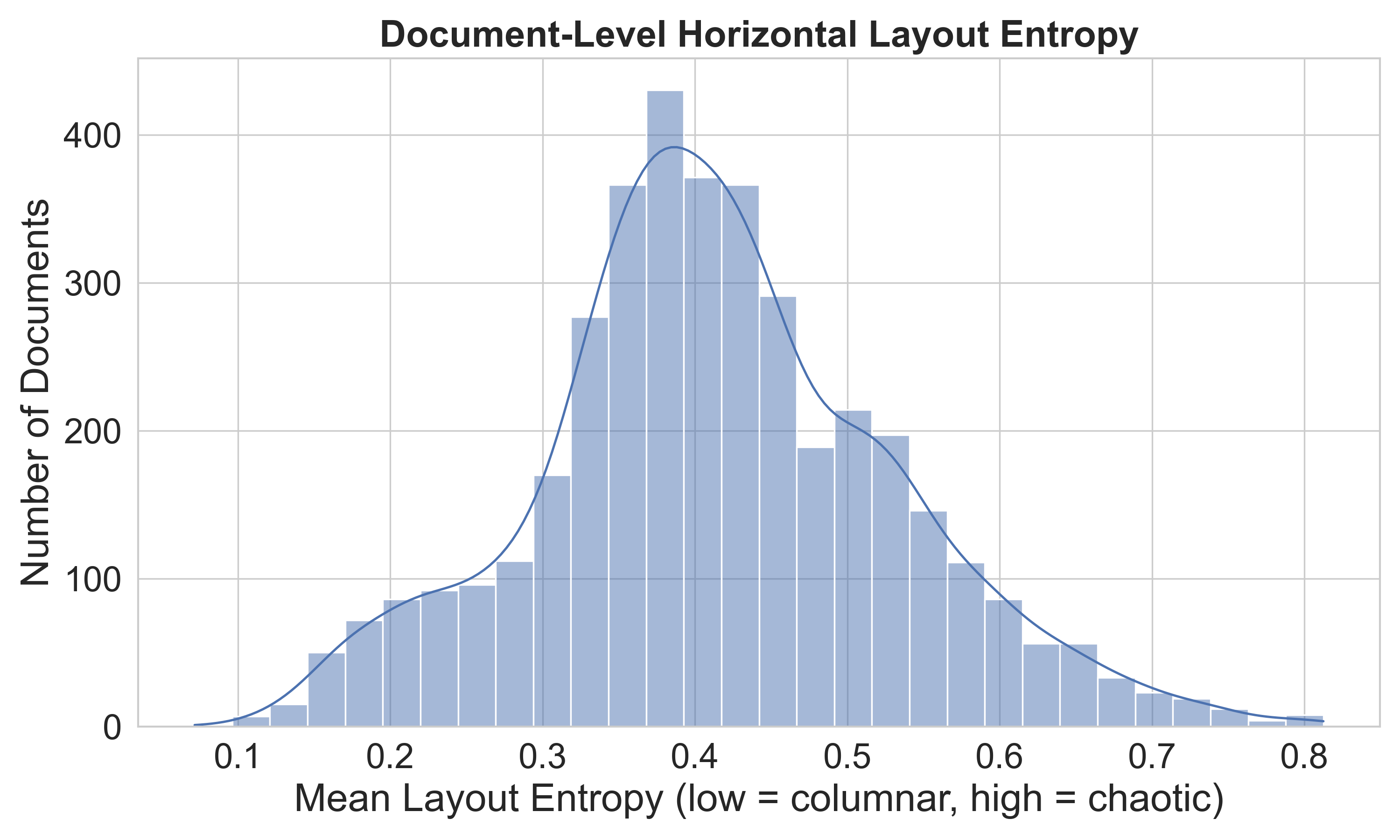}
    \caption{ Distribution of horizontal layout entropy across documents. Low entropy indicates columnar layouts with predictable vertical alignment of text blocks, while high entropy reflects chaotic layouts with irregular spatial distribution and disrupted reading order.}
    \label{fig:layout_entropy}
\end{figure}

Beyond simple entity counts, we measured the spatial organization of content using horizontal layout entropy ($H$) seen in Figure \ref{fig:layout_entropy}. This metric, calculated using Shannon entropy over soft-binned horizontal bounding box coordinates, measures the "predictability" of the document flow. Documents characterized by low entropy values ($H < 0.3$) typically represent the rigid, highly predictable columnar formats found in European Union legislative acts, where text blocks follow a strict, repetitive alignment. On the contrary, high entropy values ($H > 0.6$) indicate "chaotic" or non-linear layouts, which are prevalent in modern electronics manuals where the reading order is frequently interrupted by diagrams and multi-directional labels. By including high-entropy samples, we specifically challenge the translation system’s ability to handle fragmented text flows without losing the logical connection between spatially distant but contextually related entities.

\subsection{Granularity and Fragmentation}

\begin{figure}[ht] % Single column, [ht] tries 'here' then 'top'
\centering
    \includegraphics[width=\columnwidth]{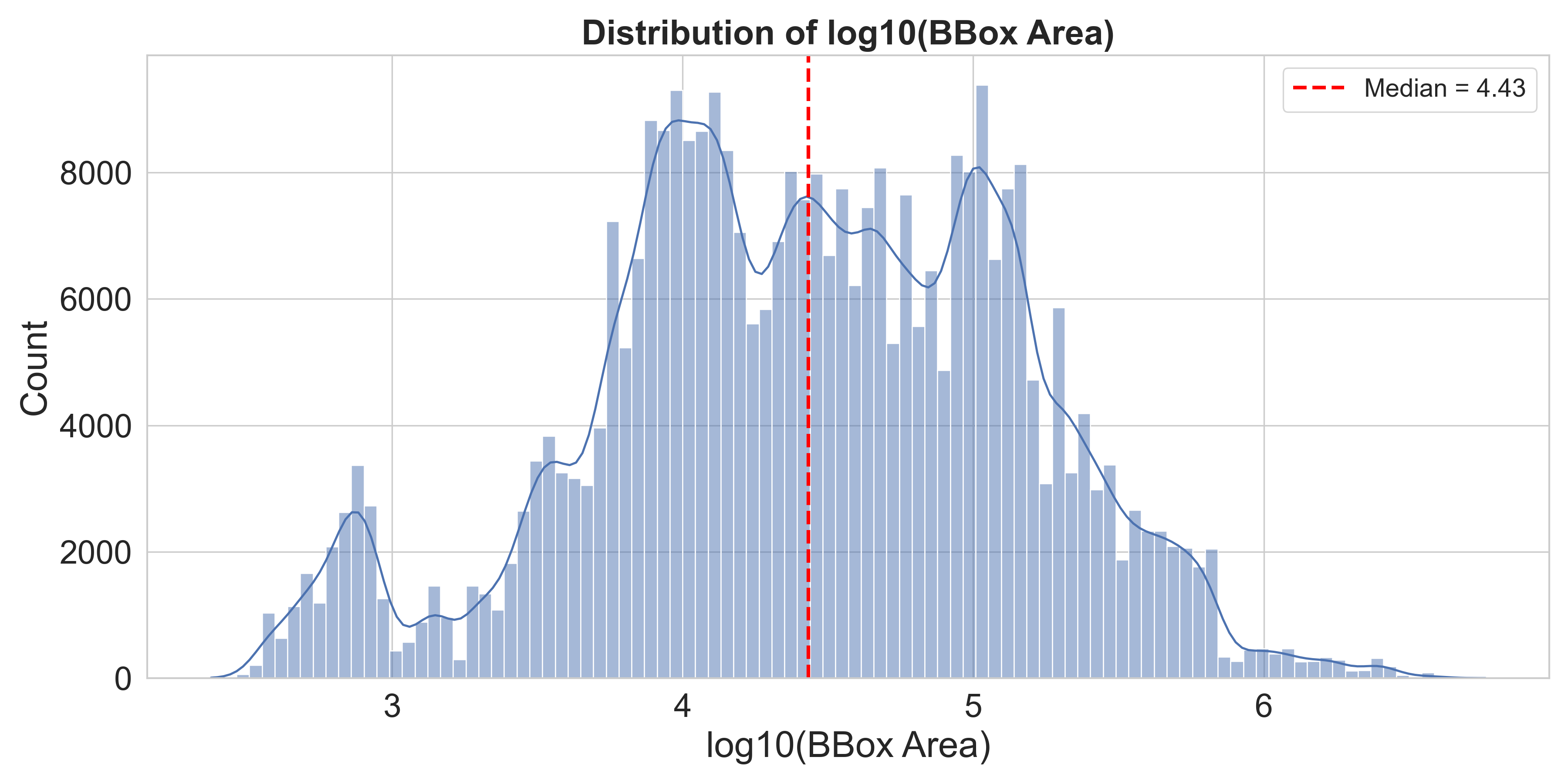}
    \caption{Bounding box area distribution on a logarithmic scale. The concentration of "micro-entities" (indicated by the peak at $\log_{10} \text{Area} \approx 4.0$) highlights the high degree of layout fragmentation.}
    \label{fig:log_bbox_area}
\end{figure}

We analyzed the physical scale of the document components. By examining the distribution of bounding box (BBox) areas on a logarithmic scale, we identified a high degree of layout fragmentation. As shown in the BBox area analysis in Figure \ref{fig:log_bbox_area}, a significant portion of the dataset consists of "micro-entities".
% ($\log_{10} \text{Area} \approx 3.0$). 
This fragmentation serves as a stress test for layout-aware translation systems, which must maintain the logical ordering and spatial coherence of these tiny, inter-dependent elements during the translation and re-rendering process.

\subsection{Spatial Density and Content Coverage}

\begin{figure}[ht] % Single column, [ht] tries 'here' then 'top'
\centering
    \includegraphics[width=\columnwidth]{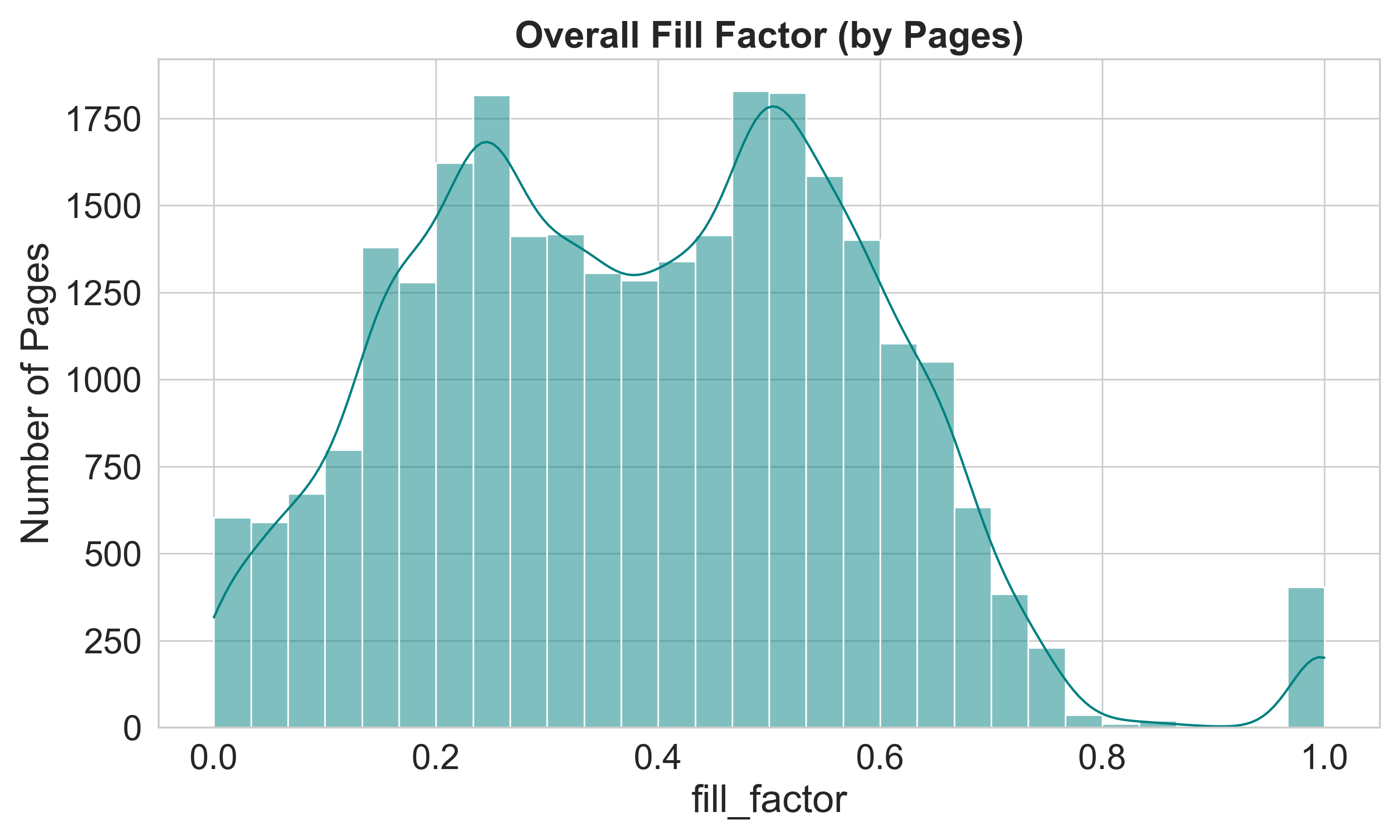}
    \caption{Distribution of the Overall Fill Factor across the corpus.}
    \label{fig:overall_fill_pages}
\end{figure}

Finally, we quantified the physical organization of the corpus using fill factor analysis, which measures the ratio of bounding box areas to the total page area. This metric provides a macroscopic view of document saturation, allowing us to categorize the corpus into distinct layout types.

As illustrated in the multi-modal distribution of Figure \ref{fig:overall_fill_pages}, the dataset captures two distinct structural profiles: low-density layouts with significant margins or white space (peaking at ~25\% coverage) and high-density pages where content saturates the layout (peaking at ~55\% coverage).

\begin{figure}[ht] % Single column, [ht] tries 'here' then 'top'
\centering
    \includegraphics[width=\columnwidth]{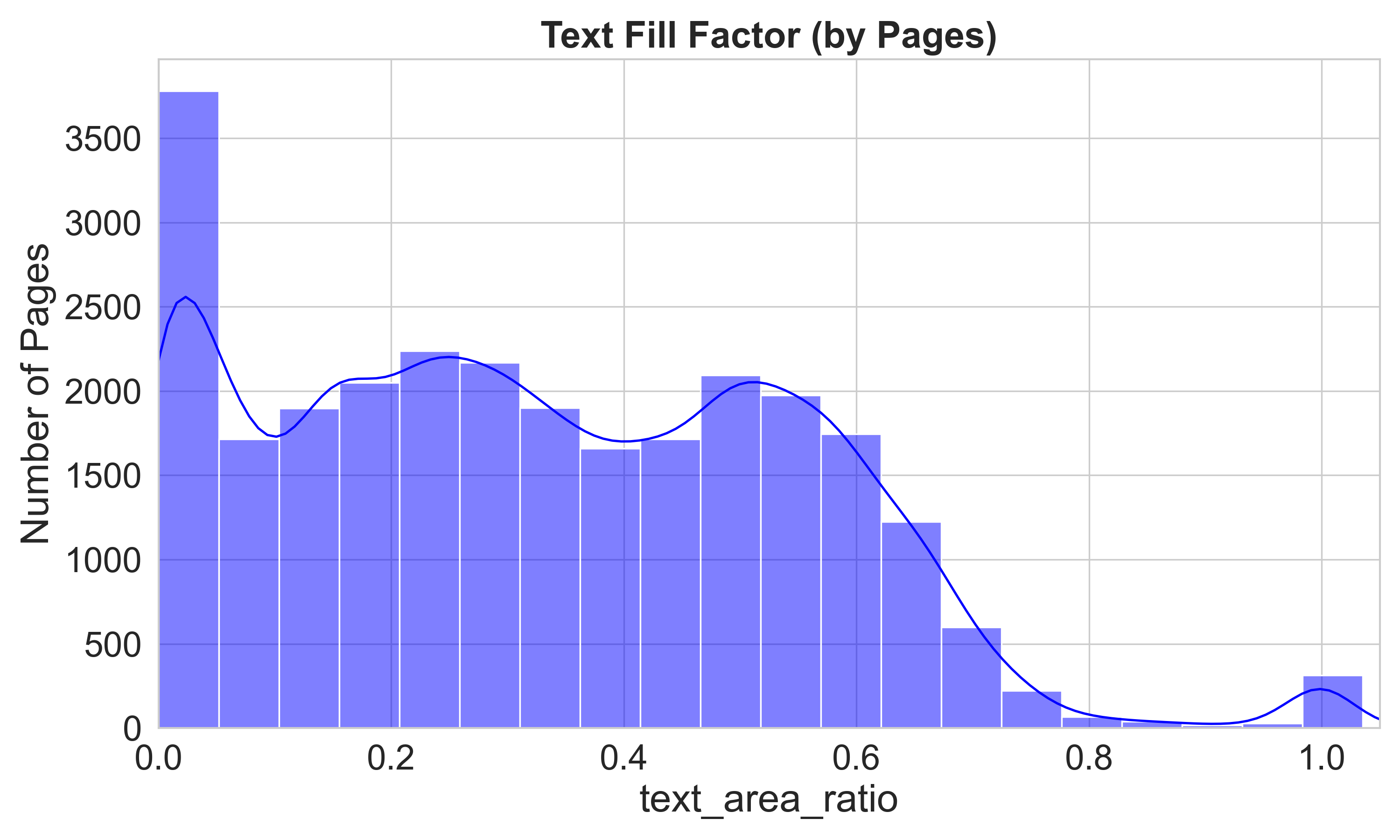}
    \caption{Text Area Ratio per page.}
    \label{fig:text_fill_pages}
\end{figure}

\begin{figure}[ht] % Single column, [ht] tries 'here' then 'top'
\centering
    \includegraphics[width=\columnwidth]{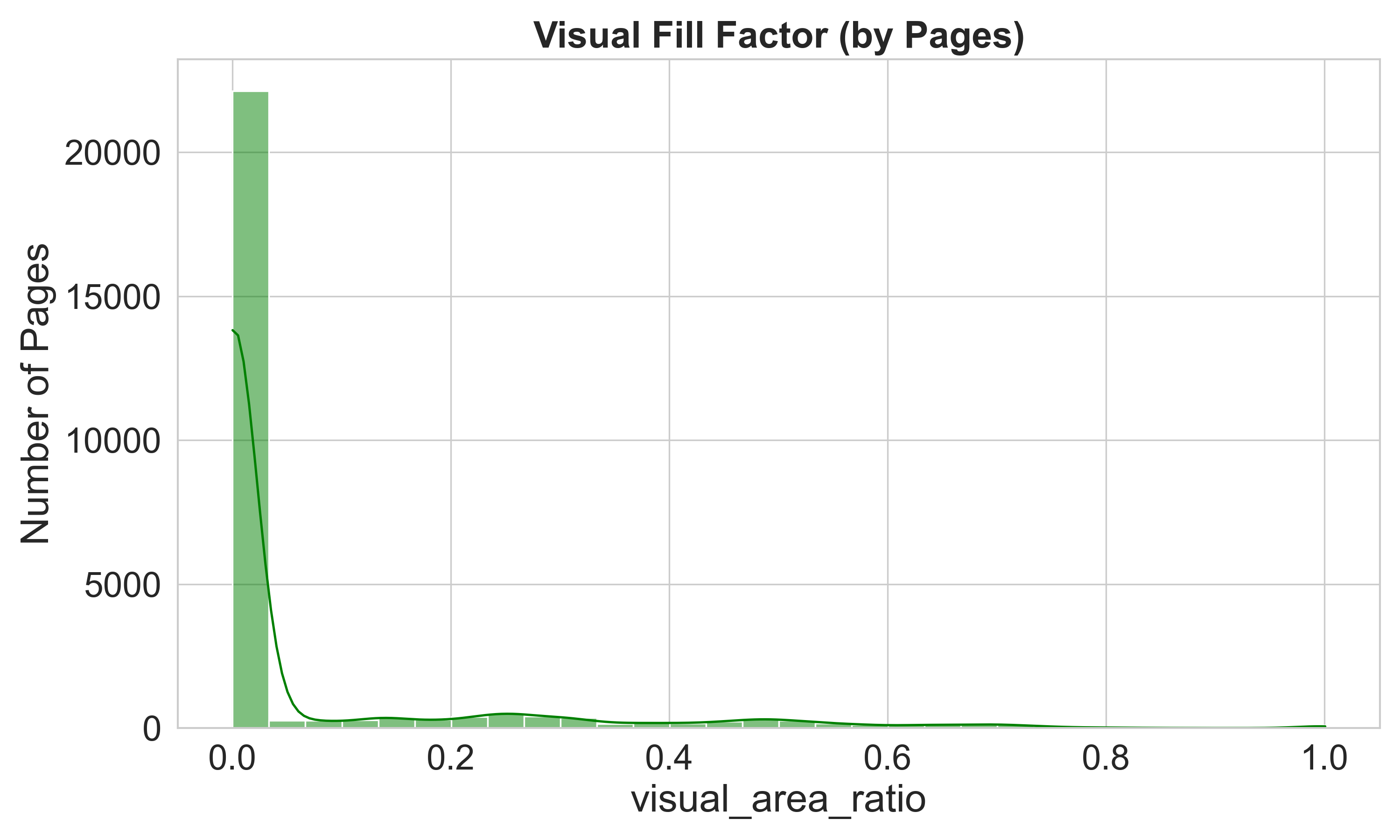}
    \caption{Visual Area Ratio per page.}
    \label{fig:visual_fill_pages}
\end{figure}

While text remains the primary content driver (Figure \ref{fig:text_fill_pages}), the Visual Fill Factor exhibits a significant long-tail distribution (Figure \ref{fig:visual_fill_pages}), with a dedicated subset of documents maintaining high structural occupancy (up to 60\% area ratio) due to the presence of large diagrams, schematics, and complex tables.

\section{Translation Systems Comparison}

To demonstrate the practical usefulness of our dataset, we performed a comparative evaluation using a manual selection of PDFs that pose distinct structural and linguistic challenges based on the documents' label counts and variety. We benchmarked 
two industry-standard commercial engines-Google Translate\footnote{\url{https://translate.google.com/}} and DeepL\footnote{\url{https://www.deepl.com/}}-as well as our internal translation system.
% our internal document translation system.
% three commercial document translation systems.
% The evaluation was conducted through a side-by-side analysis, focusing on how each system preserved the original layout and formatting in the translated output.

% \subsection{Systems Evaluated}

% which systems we chose for the translation

\subsection{Results}

% describe the results and point to issues without naming the system reponsible for given issue
% compare the systems results with the given target
%To focus on structural and linguistic challenges rather than individual system performance, we present our findings as an anonymized comparative analysis. W
We group the observed problems into two categories: linguistic errors, which stem from a system's lack of document-level context, and structural errors, which arise during the construction of the output PDF file.

\subsubsection{Translation Errors}

\begin{figure}[!ht]
    \centering
    \begin{tcolorbox}[colframe=black, boxrule=0.6pt, arc=2pt]
    \begin{subfigure}{\columnwidth}
        \centering
        \includegraphics[width=\linewidth]{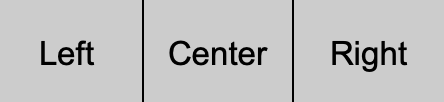}
        \caption{Source text}
    \end{subfigure}
    \vspace{0.5em}
    \begin{subfigure}{\columnwidth}
        \centering
        \includegraphics[width=\linewidth]{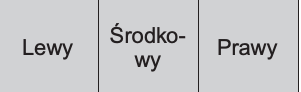}
        \caption{Original target text}
    \end{subfigure}
    \vspace{0.5em}
    \begin{subfigure}{\columnwidth}
        \centering
        \includegraphics[width=\linewidth]{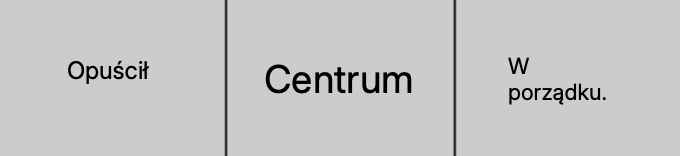}
        \caption{Translation system result}
    \end{subfigure}
    \caption{Table cells translation error presenting different semantic meaning to each cell.}
    \label{fig:trans-l-c-r}
    \end{tcolorbox}
\end{figure}

Figure \ref{fig:trans-l-c-r} illustrates a translation of three isolated table cells. In the gold-standard target text, the context refers strictly to spatial orientation: "Left," "Center," and "Right." However, one evaluated system assigned a different semantic meaning to each term. The word "Left" was mistranslated as "Opuścił" (the past tense of "to leave"), and "Right" was rendered as "W porządku" (signifying "all right" or "okay"). Furthermore, for "Center", the system opted for the noun "Centrum" instead of the required spatial adjective "Środkowy". This indicates a failure in spatial grounding, where the system lacks the table-level context.

\begin{figure}[!ht]
    \centering
    \begin{tcolorbox}[colframe=black, boxrule=0.6pt, arc=2pt]
    \begin{subfigure}[t]{0.45\columnwidth}
        \centering
        \includegraphics[width=\linewidth]{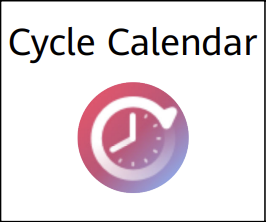}
        \caption{Source text}
    \end{subfigure}
    \hfill
    \begin{subfigure}[t]{0.45\columnwidth}
        \centering
        \includegraphics[width=\linewidth]{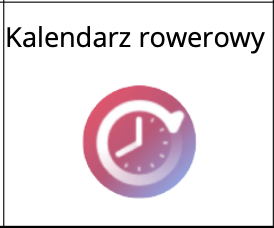}
        \caption{Translation system result}
    \end{subfigure}
    \caption{Image caption translation error introduced by missing image context.}
    \label{fig:trans-cycle}
    \end{tcolorbox}
\end{figure}

Figure \ref{fig:trans-cycle} illustrates a significant contextual dissonance in an image-caption translation task. The original document contains an icon labeled "Cycle Calendar" within a health-related context. However, the system maps "Cycle" to the biking domain, rendering the caption as "Kalendarz rowerowy" (Bicycle Calendar). This failure in multimodal grounding shows that the system prioritized common statistical associations over the actual visual context.

\subsubsection{Structural Errors}

\begin{figure}[!ht]
    \centering
    \begin{tcolorbox}[colframe=black, boxrule=0.6pt, arc=2pt]
    \begin{subfigure}{\columnwidth}
        \centering
        \includegraphics[width=\linewidth]{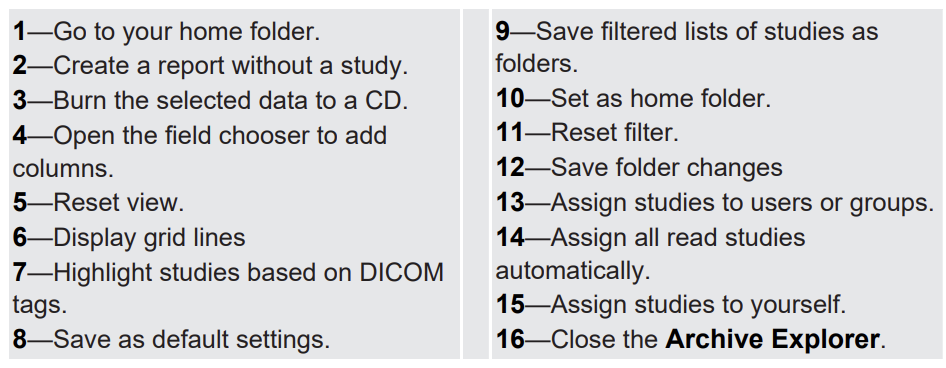}
        \caption{Source text}
    \end{subfigure}
    \vspace{0.5em}
    \begin{subfigure}{\columnwidth}
        \centering
        \includegraphics[width=\linewidth]{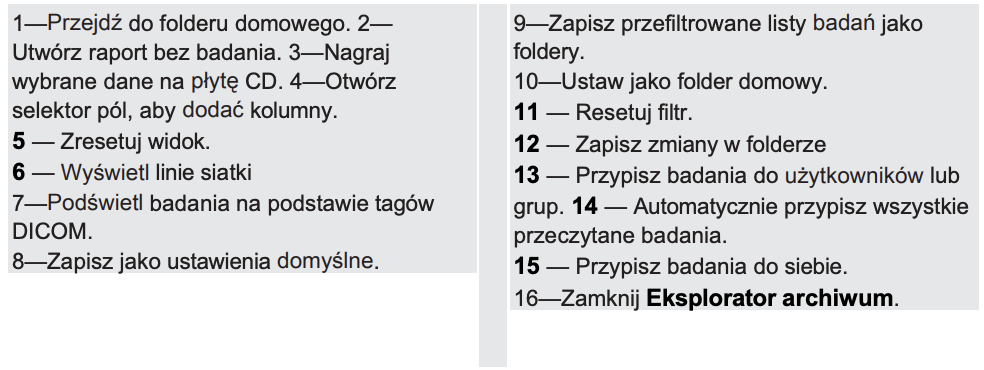}
        \caption{Translation system result}
    \end{subfigure}
    \caption{Dual-column numbered list reconstruction error. The system misplaced newline characters and resized gray background, breaking the parallel alignment.}
    \label{fig:rec-list-newlines}
    \end{tcolorbox}
\end{figure}

Figure \ref{fig:rec-list-newlines} illustrates a structural reconstruction failure in a dual-column numbered list. The evaluated system failed to preserve the original line breaks, resulting in segmentation errors where list indices were merged into preceding text blocks. Furthermore, the system recalculated the gray background box dimensions for each column independently. This lack of geometric synchronization produced uneven blocks, breaking the original typographic hierarchy and column alignment.

\begin{figure}[!ht]
    \centering
    \begin{tcolorbox}[colframe=black, boxrule=0.6pt, arc=2pt]
    \begin{subfigure}{\columnwidth}
        \centering
        \includegraphics[width=\linewidth]{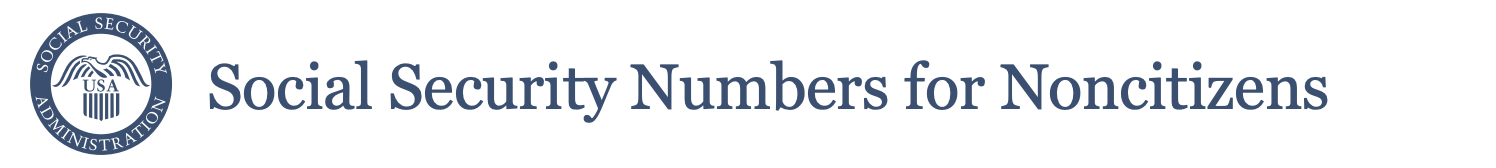}
        \caption{Source text}
    \end{subfigure}
    \vspace{0.5em}
    \begin{subfigure}{\columnwidth}
        \centering
        \includegraphics[width=\linewidth]{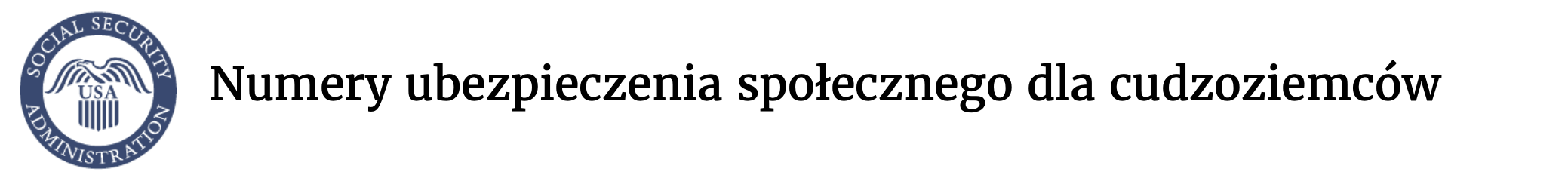}
        \caption{Translation system result}
    \end{subfigure}
    \begin{subfigure}{\columnwidth}
        \centering
        \includegraphics[width=\linewidth]{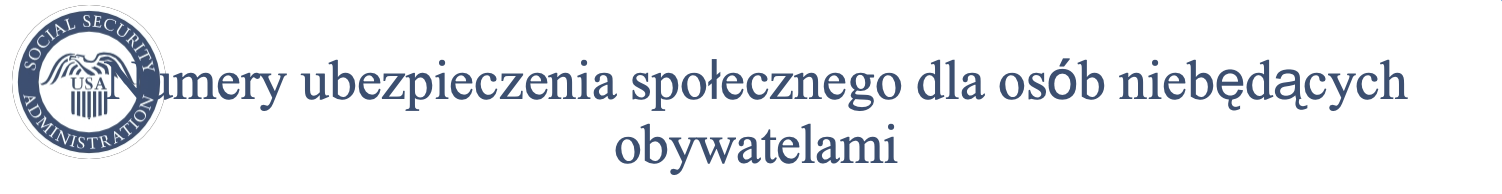}
        \caption{Translation system result}
    \end{subfigure}
    \caption{Document header reconstruction errors. One system failed to restore blue text coloring, while the second one overlayed the text over a logo and rendered some letters in different typeface.}
    \label{fig:rec-social}
    \end{tcolorbox}
\end{figure}

Figure \ref{fig:rec-social} displays a header featuring the "Social Security Administration" logo with adjacent blue text. The evaluated systems exhibited distinct reconstruction errors. The first system failed to preserve the text color metadata, rendering the text in a default black font. The second system produced a collision, overlaying the translated text partially onto the logo. Additionally, this system suffered from font-fallback artifacts, where the Latin-1 characters (English) and the extended Latin characters (Polish diacritics) were rendered in mismatched typefaces.

\begin{table}[ht]
\centering
\small
\caption{System performance across structural and semantic challenges. We compared three systems: Ours: our internal translator; DeepL and GoogleT: Google Translate).}
\label{tab:structural_benchmarking}
\begin{tabularx}{\columnwidth}{@{}l X ccc @{}}
\toprule
\textbf{Category} & \textbf{Structural Feature} & \textbf{Ours} & \textbf{DeepL} & \textbf{GoogleT} \\ 
\midrule
\textbf{Trans.} & \textbf{Semantic Inversion} \newline (Table Cells) & \textit{Fail} & \textit{Pass} & \textit{Partial} \\ 
\addlinespace
\textbf{Trans.} & \textbf{Multimodal Grounding} \newline (Image Captions) & \textit{Fail} & \textit{Pass} & \textit{Fail} \\ 
\midrule
\textbf{Layout} & \textbf{Metadata Preservation} \newline (Color/Position) & \textit{Partial} & \textit{Partial} & \textit{Pass} \\ 
\addlinespace
\textbf{Layout} & \textbf{Lists Reconstruction} \newline (Dual-Columns) & \textit{Pass} & \textit{Fail} & \textit{Partial} \\ 
\bottomrule
\end{tabularx}
\end{table}

The observed linguistic and structural limitations are synthesized in Table \ref{tab:structural_benchmarking}, which provides an overview of how each system managed the core challenges identified in our corpus.

\section{Conclusion}

In this work, we introduced a novel dataset comprising 3,956 PDF documents across 15 language pairs, sourced from the legal and technical domains. A defining characteristic of this corpus is its high layout and formatting complexity, which is essential for evaluating end-to-end document translation pipelines. By utilizing a hybrid extraction pipeline and K-Medoids clustering over a set of 45 features, we ensured the dataset captures a diverse set of document layouts.

In contrast to conventional plain-text datasets commonly used in machine translation, this benchmark preserves the full layout and typographic context of the original documents, thereby providing a more accurate representation of real-world translation scenarios. This addresses a critical gap in machine translation research, where visual context is frequently discarded or limited to a single reference image.

Our qualitative analysis of commercial translation systems highlights significant limitations in current architectures when processing visually-rich documents. The evaluated systems frequently fail to maintain structural and formatting integrity during the translation process.

This dataset serves as a benchmark for evaluating layout-aware and document-level translation systems. We expect it to drive future research toward models that jointly optimize for linguistic accuracy, contextual translation, and formatting preservation. We leave the development of automatic metrics for evaluating formatting preservation to future work.

\section{Limitations}
The released dataset is limited to left-to-right languages written in the Latin alphabet as we focus on main European languages. Moreover, long documents are limited to 10 pages only as the goal of this benchmark is to focus on visual context, which frequently is quite local.

%Although documents in other writing systems were considered during collection, they were excluded from the final release. 
%The dataset therefore does not cover evaluation on non-Latin or non-left-to-right writing systems.

% \bibliography{\confname}

\bibliography{eamt23}
\bibliographystyle{eamt23}

% \clearpage

% \section*{Appendix A. Vector Indices Mapping}

\begin{table*}[t] % The asterisk * makes it span both columns
\centering
\section*{Appendix A. Vector Indices Mapping}
\small
\setlength{\tabcolsep}{8pt} % Adjusts horizontal spacing between columns
\caption{Full enumeration of the 45-dimensional document feature vector mapping.}
\label{tab:vector_mapping}
\begin{tabular}{ll | ll | ll}
\hline
\textbf{Idx} & \textbf{Entity / Attribute} & \textbf{Idx} & \textbf{Entity / Attribute} & \textbf{Idx} & \textbf{Entity / Attribute} \\ \hline
0 & \texttt{algorithm} & 15 & \texttt{footer\_image} & 30 & \texttt{diff\_font\_names} \\
1 & \texttt{ref\_content} & 16 & \texttt{text} & 31 & \texttt{diff\_font\_sizes} \\
2 & \texttt{doc\_title} & 17 & \texttt{number} & 32 & \texttt{black} \\
3 & \texttt{vision\_footnote} & 18 & \texttt{chart} & 33 & \texttt{white} \\
4 & \texttt{inline\_formula} & 19 & \texttt{header\_image} & 34 & \texttt{gray} \\
5 & \texttt{display\_formula} & 20 & \texttt{image} & 35 & \texttt{pink} \\
6 & \texttt{footer} & 21 & \texttt{abstract} & 36 & \texttt{beige} \\
7 & \texttt{reference} & 22 & \texttt{content} & 37 & \texttt{brown} \\
8 & \texttt{seal} & 23 & \texttt{paragraph\_title} & 38 & \texttt{red} \\
9 & \texttt{table} & 24 & \texttt{figure\_title} & 39 & \texttt{orange} \\
10 & \texttt{aside\_text} & 25 & \texttt{vertical} & 40 & \texttt{yellow} \\
11 & \texttt{formula\_num} & 26 & \texttt{normal} (weight) & 41 & \texttt{green} \\
12 & \texttt{footnote} & 27 & \texttt{bold} (weight) & 42 & \texttt{teal\_cyan} \\
13 & \texttt{vertical\_text} & 28 & \texttt{italic} (weight) & 43 & \texttt{blue} \\
14 & \texttt{header} & 29 & \texttt{bolditalic} (weight) & 44 & \texttt{purple} \\ \hline
\end{tabular}
\end{table*}

\clearpage
% \section*{Appendix B. Clustering Impact}

\begin{table*}
\centering
\section*{Appendix B. Clustering Impact}
\phantomsection
\label{appendix:b:impact}
\vspace{0.5cm}
\caption{Impact of the clustering methodology on typographic color distribution.}
\begin{tabular}{lrrr}
\hline
Color & Before (\%) & After (\%) & Change (\%) \\
\hline
Yellow & 0.00062743 & 0.00228579 & +264.31\% \\
Red & 0.01930891 & 0.06094932 & +215.65\% \\
Purple & 0.01514131 & 0.03878174 & +156.13\% \\
White & 0.774143 & 1.74502507 & +125.41\% \\
Teal\_Cyan & 0.0067147 & 0.01469656 & +118.87\% \\
Pink & 4.986e-05 & 0.00010739 & +115.38\% \\
Orange & 0.0006856 & 0.0014267 & +108.10\% \\
Green & 0.19981624 & 0.38189586 & +91.12\% \\
Black & 96.60691349 & 96.30528719 & -0.31\% \\
Gray & 1.92919544 & 1.21493631 & -37.02\% \\
Blue & 0.44740404 & 0.23460807 & -47.56\% \\
\hline
\end{tabular}

% \end{table}

\vspace{1cm}

% \begin{table}[ht]
\centering
\caption{Impact of the clustering methodology on entity labels distribution.}
\begin{tabular}{lrrr}
\hline
Label & Before (\%) & After (\%) & Change (\%) \\
\hline
aside\_text & 0.65733778 & 1.58451011 & +141.05\% \\
inline\_formula & 0.02724798 & 0.05804066 & +113.01\% \\
content & 2.44546664 & 5.1188963 & +109.32\% \\
image & 2.06501872 & 4.15455068 & +101.19\% \\
table & 0.96620069 & 1.55432896 & +60.87\% \\
vision\_footnote & 0.20388733 & 0.30239186 & +48.31\% \\
chart & 0.00834765 & 0.01160813 & +39.06\% \\
figure\_title & 0.14120123 & 0.18776155 & +32.97\% \\
paragraph\_title & 12.50579807 & 14.15292554 & +13.17\% \\
number & 5.94651679 & 6.30495725 & +6.03\% \\
footer & 5.25318963 & 4.90675767 & -6.59\% \\
reference & 0.00094502 & 0.00087061 & -7.87\% \\
header & 5.69340985 & 5.24455433 & -7.88\% \\
text & 57.56836211 & 51.29604801 & -10.90\% \\
header\_image & 1.11582832 & 0.9536081 & -14.54\% \\
reference\_content & 0.19420091 & 0.15670979 & -19.31\% \\
footnote & 2.23606672 & 1.76472637 & -21.08\% \\
doc\_title & 1.21261377 & 0.93938814 & -22.53\% \\
footer\_image & 1.63763497 & 1.25483914 & -23.37\% \\
seal & 0.00039376 & 0.0002902 & -26.30\% \\
abstract & 0.08284645 & 0.04643253 & -43.95\% \\
formula\_number & 0.00102377 & 0.0002902 & -71.65\% \\
algorithm & 0.02661797 & 0.00406285 & -84.74\% \\
display\_formula & 0.00984392 & 0.00145102 & -85.26\% \\
\hline
\end{tabular}
\end{table*}

\clearpage
\newpage
\section*{Appendix C. Additional Translation Errors Examples}

\begin{figure}[!ht]
    \centering
    \begin{tcolorbox}[colframe=black, boxrule=0.6pt, arc=2pt]
    \begin{subfigure}{\columnwidth}
        \centering
        \includegraphics[width=\linewidth]{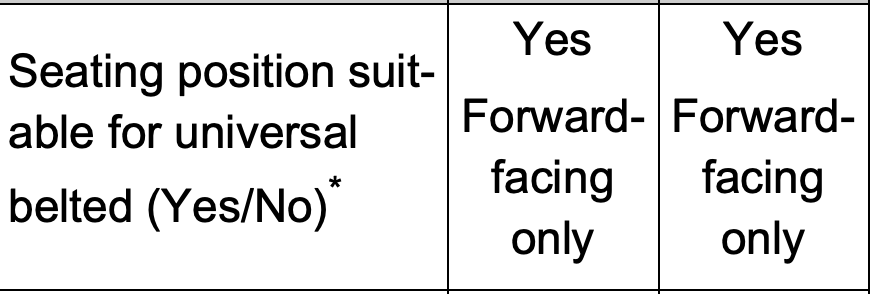}
        \caption{Source text}
    \end{subfigure}
    \vspace{0.5em}
    \begin{subfigure}{\columnwidth}
        \centering
        \includegraphics[width=\linewidth]{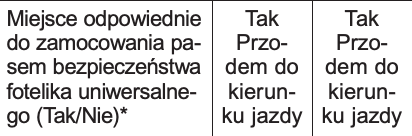}
        \caption{Original target text}
    \end{subfigure}
    \vspace{0.5em}
    \begin{subfigure}{\columnwidth}
        \centering
        \includegraphics[width=\linewidth]{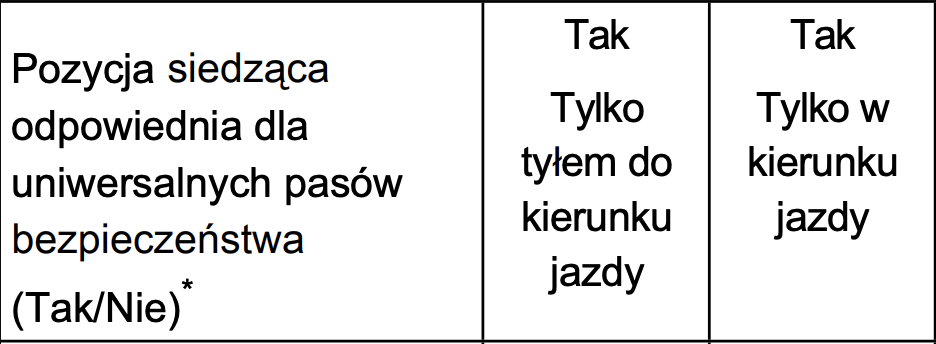}
        \caption{Translation system result}
    \end{subfigure}
    \caption{(Table cells translation error. In the center column, the system produces a critical semantic inversion, incorrectly translating the instruction as "Tylko tyłem" (Rear-facing only). This error represents a significant user safety risk and demonstrates the danger of translating table cells without structural context.}
    \end{tcolorbox}
\end{figure}

\begin{figure}[!ht]
    \centering
    \begin{tcolorbox}[colframe=black, boxrule=0.6pt, arc=2pt]
    \begin{subfigure}{\columnwidth}
        \centering
        \includegraphics[width=\linewidth]{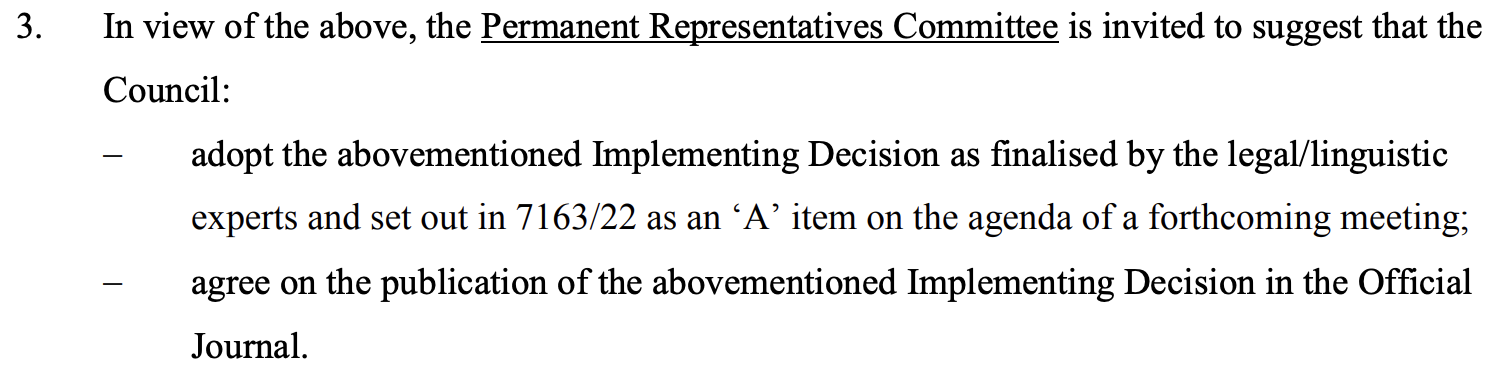}
        \caption{Source text}
    \end{subfigure}
    \vspace{0.5em}
    \begin{subfigure}{\columnwidth}
        \centering
        \includegraphics[width=\linewidth]{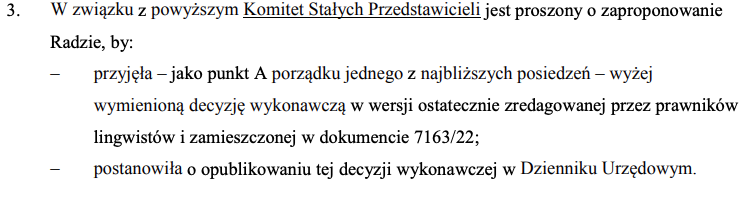}
        \caption{Original target text}
    \end{subfigure}
    \vspace{0.5em}
    \begin{subfigure}{\columnwidth}
        \centering
        \includegraphics[width=\linewidth]{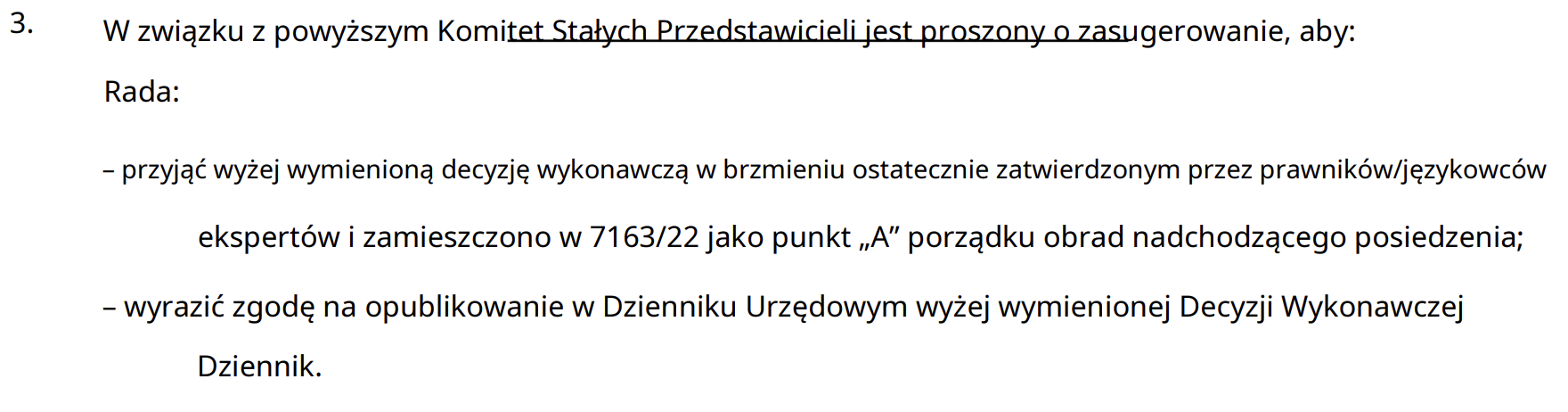}
        \caption{Translation system result}
    \end{subfigure}
    \caption{Translation system losing semantic translation context between lines and misplacing the text underline.}
    \end{tcolorbox}
\end{figure}

\begin{figure}[!ht]
    \centering
    \begin{tcolorbox}[colframe=black, boxrule=0.6pt, arc=2pt]
    \begin{subfigure}{\columnwidth}
        \centering
        \includegraphics[width=\linewidth]{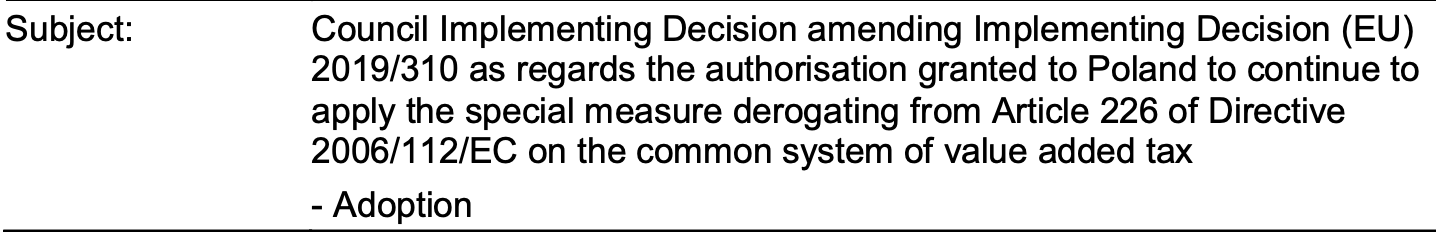}
        \caption{Source text}
    \end{subfigure}
    \vspace{0.5em}
    \begin{subfigure}{\columnwidth}
        \centering
        \includegraphics[width=\linewidth]{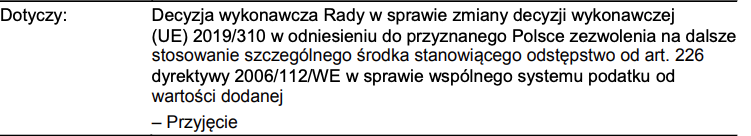}
        \caption{Original target text}
    \end{subfigure}
    \vspace{0.5em}
    \begin{subfigure}{\columnwidth}
        \centering
        \includegraphics[width=\linewidth]{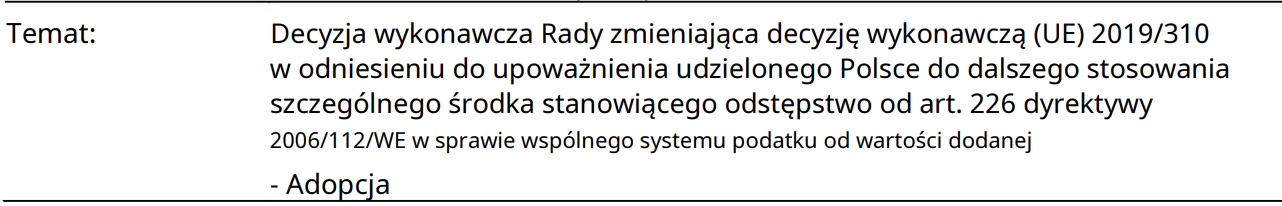}
        \caption{Translation system result}
    \end{subfigure}
    \begin{subfigure}{\columnwidth}
        \centering
        \includegraphics[width=\linewidth]{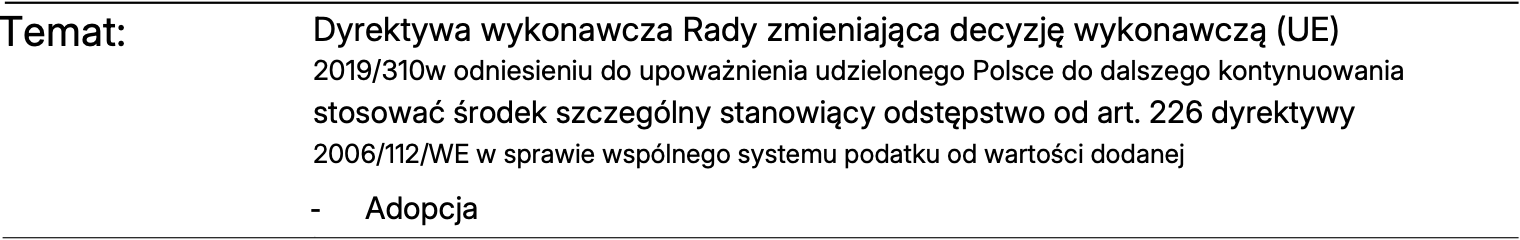}
        \caption{Translation system result}
    \end{subfigure}
    \caption{The ground-truth translation correctly renders "Adoption" as "Przyjęcie" (Legal Adoption/Approval) to match the legislative subject matter. However, the tested systems exhibit a significant domain mismatch, mistranslating the term as "Adopcja" (Biological/Family Adoption). This error stems from a loss of contextual continuity between layout elements.}
    \end{tcolorbox}
\end{figure}

% \begin{figure}[H]
%     \centering
%     \begin{subfigure}[t]{0.45\columnwidth}
%         \centering
%         \includegraphics[width=\linewidth]{images/translation_errors/translation/6/og.png}
%         \caption{Source text}
%     \end{subfigure}
%     \hfill
%     \begin{subfigure}[t]{0.45\columnwidth}
%         \centering
%         \includegraphics[width=\linewidth]{images/translation_errors/translation/6/google.png}
%         \caption{Translation system result}
%     \end{subfigure}
%     \caption{The translation system incorrectly renders the noun "Workout" as the infinitive verb "Ćwiczyć" (to exercise), whereas the visually indicated context of a fitness application requires the noun "Trening".}
% \end{figure}

\clearpage
% \newpage

\section*{Appendix D. Additional Reconstruction Errors Examples}

\begin{figure}[!ht]
    \centering
    \begin{tcolorbox}[colframe=black, boxrule=0.6pt, arc=2pt]
    \begin{subfigure}{\columnwidth}
        \centering
        \includegraphics[width=\linewidth]{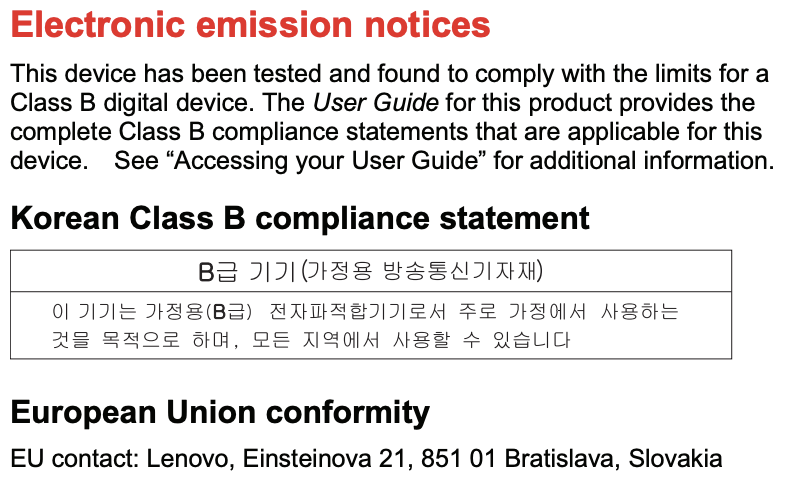}
        \caption{Source text}
    \end{subfigure}
    \vspace{0.5em}
    \begin{subfigure}{\columnwidth}
        \centering
        \includegraphics[width=\linewidth]{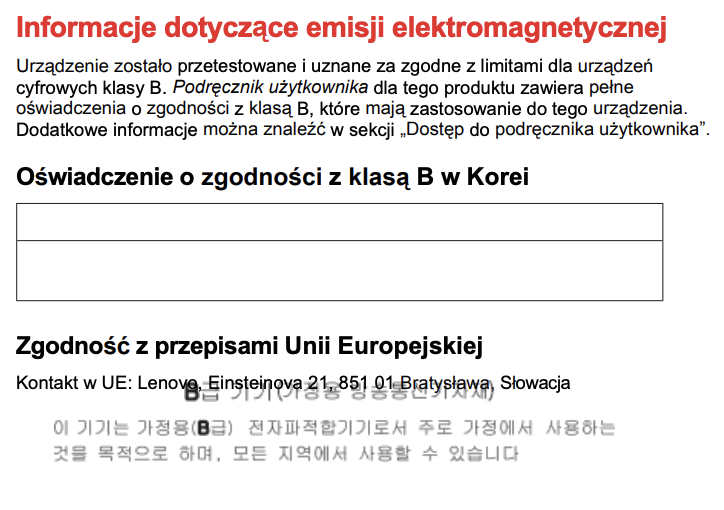}
        \caption{Translation system result}
    \end{subfigure}
    \caption{Geometric synchronization failure and layer detachment. In the original document, the compliance statement is properly encapsulated within a table structure. However, the translation system fails to maintain the link between the bounding box and its textual content.}
    \end{tcolorbox}
\end{figure}

% \begin{figure}[!ht]
%     \centering
%     \begin{subfigure}{\columnwidth}
%         \centering
%         \includegraphics[width=\linewidth]{images/translation_errors/reconstruction/3/og.png}
%         \caption{Source text}
%     \end{subfigure}
%     \vspace{0.5em}
%     \begin{subfigure}{\columnwidth}
%         \centering
%         \includegraphics[width=\linewidth]{images/translation_errors/reconstruction/3/laniqo.png}
%         \caption{Translation system result}
%     \end{subfigure}
%     \caption{Two images side by side}
% \end{figure}

\begin{figure}[!ht]
    \centering
    \begin{tcolorbox}[colframe=black, boxrule=0.6pt, arc=2pt]
    \begin{subfigure}{\columnwidth}
        \centering
        \includegraphics[width=\linewidth]{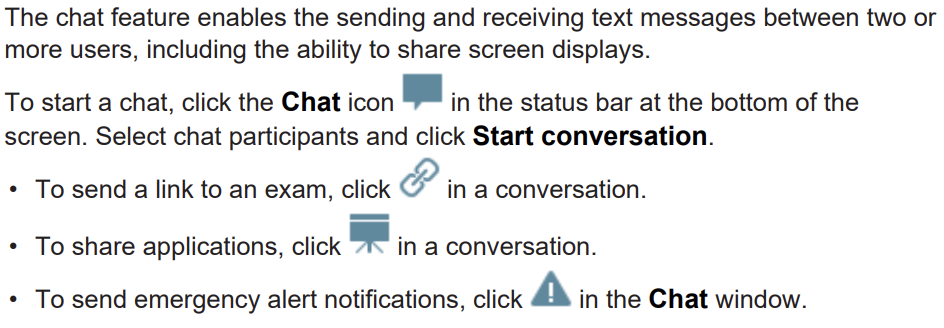}
        \caption{Source text}
    \end{subfigure}
    \vspace{0.5em}
    \begin{subfigure}{\columnwidth}
        \centering
        \includegraphics[width=\linewidth]{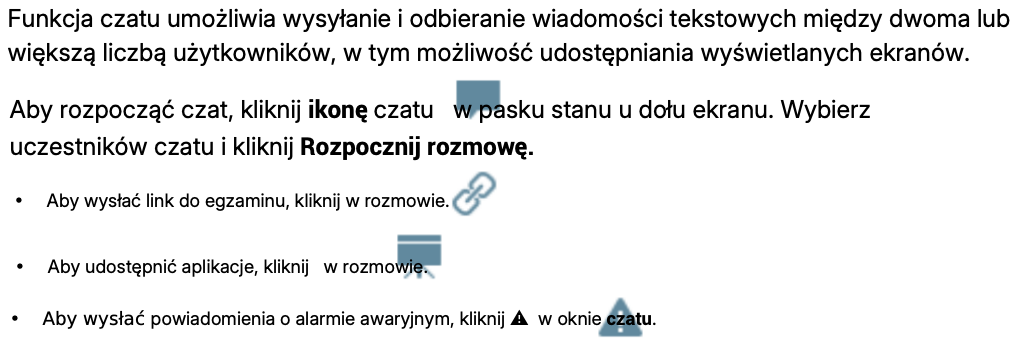}
        \caption{Translation system result}
    \end{subfigure}
    \caption{Inline asset collision and anchor failure. The original document features functional icons embedded within the text flow. The translation system fails to account for these inline graphical assets during the reconstruction phase.}
    \end{tcolorbox}
\end{figure}

\begin{figure}[!ht]
    \centering
    \begin{tcolorbox}[colframe=black, boxrule=0.6pt, arc=2pt]
    \begin{subfigure}[t]{0.45\columnwidth}
        \centering
        \includegraphics[width=\linewidth]{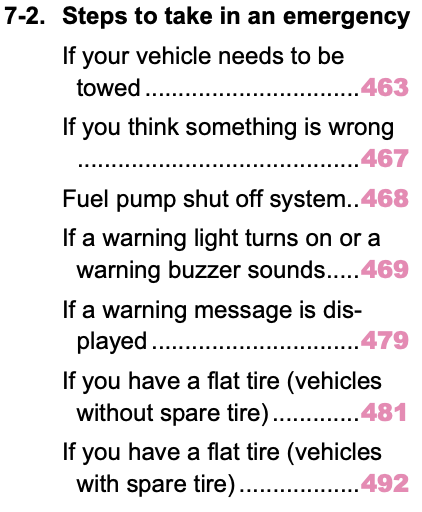}
        \caption{Source text}
    \end{subfigure}
    \hfill
    \begin{subfigure}[t]{0.45\columnwidth}
        \centering
        \includegraphics[width=\linewidth]{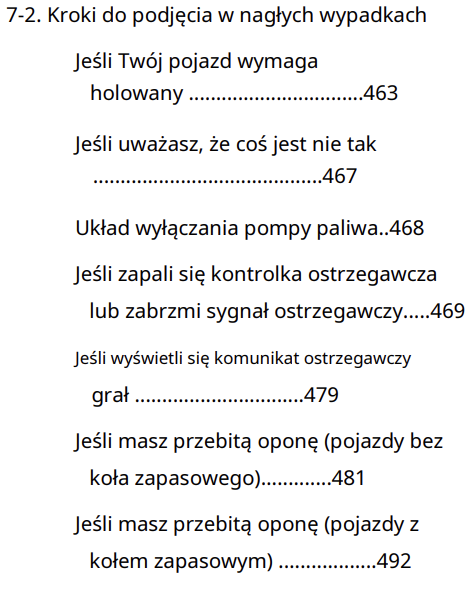}
        \caption{Translation system result}
    \end{subfigure}
    \caption{Failure in structural reconstruction and stylistic preservation. In the reconstructed output, the system fails to preserve the typographic weight and the pink-colored indices, rendering all elements in a default black font. Furthermore, the translation exhibits a significant vertical alignment drift.}
    \end{tcolorbox}
\end{figure}

\end{document}